\PassOptionsToPackage{table}{xcolor}
\documentclass[letterpaper]{article}
\usepackage[preprint]{aaai2027}
\usepackage[hyphens]{url}
\usepackage{graphicx}
\usepackage{amsmath}
\usepackage{natbib}
\usepackage{caption}
\usepackage{booktabs}
\usepackage{amssymb}
\usepackage{multirow}
\usepackage{tabularx}
\definecolor{ReferenceRow}{HTML}{F4F5F5}
\definecolor{OursRow}{HTML}{EAF5F1}
\newcommand{\bestresult}[1]{\textbf{#1}}
\newcommand{\secondresult}[1]{\underline{#1}}
\title{Recall Before You Rank: Similarity-Guided Top-$K$ Reuse for Efficient Long-Context Attention}
\author{
Wenshuai Yao\textsuperscript{\rm 1}\equalcontrib,
Wenyong Zhou\textsuperscript{\rm 2}\equalcontrib,
Hanyong Shao\textsuperscript{\rm 1},
Yizhe Chen\textsuperscript{\rm 3},\\
Zhiyuan Ning\textsuperscript{\rm 1},
Yuannuo Feng\textsuperscript{\rm 3},
Ru Huang\textsuperscript{\rm 1}\corresponding,
Kechao Tang\textsuperscript{\rm 1}\corresponding
}
\affiliations{
\textsuperscript{\rm 1}School of Integrated Circuits, Peking University, Beijing, China\\
\textsuperscript{\rm 2}Department of Electrical and Computer Engineering, The University of Hong Kong, Hong Kong SAR, China\\
\textsuperscript{\rm 3}School of Integrated Circuit Science and Engineering, Beihang University, Beijing, China
}

\begin{document}

\maketitle

\begin{abstract}
Top-$K$ sparse attention reduces the cost of Softmax and value aggregation by attending to only a small subset of key--value (KV) entries. However, identifying this subset still requires scoring the current query against the full KV cache and performing global Top-$K$ selection, leaving selector cost linear in context length and limiting the practical efficiency of sparse attention for long-context decoding.
In this paper, we introduce \textbf{ReTopK}, a training-free method that accelerates dynamic Top-$K$ attention by reusing historical retrieval decisions. ReTopK builds on the observation that similar queries often attend to overlapping supports and that partially overlapping supports can still preserve most of the Exact Top-$K$ attention mass. For each attention head, it maintains a bounded cache of historical query--support pairs, retrieves the most similar cached queries for each new query, unions their stored supports with a recent window, and reranks only the resulting compact candidate set using exact current-query scores. A similarity-based fallback invokes full-history Exact Top-$K$ when reuse is unreliable, while periodic exact refreshes limit cache drift. ReTopK retains the complete KV cache and reuses only selected indices, rather than historical scores, attention weights, or outputs.
Across 16K--128K contexts, ReTopK achieves the lowest PG19 perplexity and the highest NIAH and LongBench scores among the evaluated approximate methods. At 128K with $K=512$, ReTopK incurs only a 0.50\% perplexity increase over Exact Top-$K$ while accelerating attention computation by $3.07\times$.
\end{abstract}

\section{Introduction}

Large language models (LLMs) increasingly process long documents, code repositories, and extended interaction histories. During autoregressive decoding, however, the cost of attention grows with the key--value (KV) cache: each new query must compare against a growing history of keys and aggregate the corresponding values. Although optimized kernels such as FlashAttention~\citep{dao2022flashattention,dao2023flashattention2} improve hardware efficiency, they do not remove the linear dependence of decoding attention on context length.

\begin{figure}[!t]
\centering
\includegraphics[width=\columnwidth]{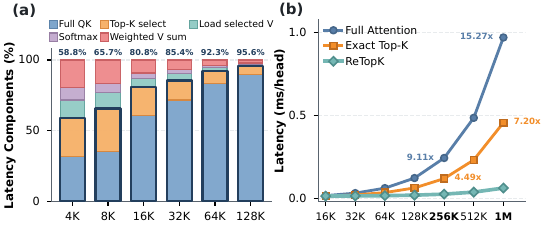}
\caption{Long-context attention latency: (a) Exact Top-$K$ breakdown at
$K=512$, with full-history QK and Top-$K$ selection outlined; (b) per-head
BF16 latency.}
\label{fig:teaser}
\end{figure}

Top-$K$ sparse attention offers a promising alternative by restricting Softmax and value aggregation to the $K$ highest-scoring KV entries. Yet, conventional Exact Top-$K$ attention must still score the current query against the entire KV cache and perform global Top-$K$ selection before sparse aggregation can begin. Thus, it sparsifies attention aggregation but not index discovery. As shown in Figure~\ref{fig:teaser}, full-history query--key (QK) scoring and Top-$K$ selection dominate the latency of Exact Top-$K$ attention, accounting for an increasingly large fraction of runtime as the context grows. Consequently, efficient long-context sparse attention requires reducing not only the cost of attending to selected entries, but also the cost of identifying them.

Existing approaches address this challenge from complementary directions. KV-cache reduction and streaming methods bound attention cost by retaining a limited subset of tokens~\citep{xiao2024streamingllm,zhang2023h2o}, but discarded positions are unavailable if they become relevant later. Query-aware methods preserve more dynamic access to the context through page-level pruning, compressed representations, or approximate retrieval~\citep{tang2024quest,singhania2024loki}; however, their selection procedures typically still inspect a search space that grows with context length. This leaves an important question: \emph{can a dynamic selector recover query-specific attention supports without searching the full history at every decoding step?}

Our answer is motivated by a simple observation: similar queries often induce overlapping Top-$K$ supports, and even incomplete support overlap can preserve most of the attention mass. Historical query--support pairs therefore provide a compact record of prior retrieval decisions. Rather than recomputing a full-history Top-$K$ support for every query, the selector can first recall supports associated with similar historical queries and then rerank only their union.

In this paper, we introduce \textbf{ReTopK}, a training-free method that accelerates dynamic Top-$K$ attention through retrieval-decision reuse. ReTopK maintains a bounded query--support cache independently for each attention head. For a new query, it retrieves the most similar cached queries, unions their stored Top-$K$ supports with a recent-token window, and reranks this compact candidate set using exact current-query scores. The resulting support is then used for sparse Softmax--value aggregation. ReTopK uses historical index selections only as candidates: it does not reuse historical attention scores, weights, or outputs. To avoid unreliable reuse, it falls back to full-history Exact Top-$K$ when cached-query similarity is low and periodically refreshes the cache with exact supports.
Our contributions are as follows:
\begin{itemize}
    \item We identify full-history index discovery as a central bottleneck of Exact Top-$K$ attention and show that query similarity provides an effective signal for reusing sparse attention supports.
    \item We propose ReTopK, a training-free recall-before-rerank method that combines a per-head query--support cache, similarity-guided candidate construction, exact candidate reranking, similarity-based fallback, and periodic refresh.
    \item We develop a fused GPU implementation and evaluate ReTopK across multiple models and long-context tasks. At 128K with $K=512$, ReTopK incurs only a 0.50\% perplexity increase over Exact Top-$K$ while accelerating attention computation by $3.07\times$.
\end{itemize}

\section{Related Work}

\paragraph{Efficient attention and Exact Top-$K$ attention.}
IO-aware attention kernels, including FlashAttention and FlashAttention-2, improve the efficiency of dense attention through tiling, fusion, and reduced memory traffic~\cite{dao2022flashattention,dao2023flashattention2}. However, autoregressive decoding still requires each query to interact with a KV cache whose length grows with the context. Exact Top-$K$ attention reduces the cost of Softmax--value aggregation by retaining only the globally highest-scoring $K$ entries, but it must first compute QK scores over the full history and perform global Top-$K$ selection. ReTopK targets this remaining index-discovery cost: it approximates the support-selection process while retaining exact current-query scoring within the recalled candidate set.

\paragraph{KV-cache reduction and structured sparsity.}
A complementary line of work bounds attention cost by reducing or organizing the active KV cache. Streaming and eviction methods retain attention sinks, recent tokens, or tokens selected by accumulated importance~\cite{xiao2024streamingllm,liu2023scissorhands,zhang2023h2o}. Other approaches compress the cache using observation-window signals, layer-dependent budgets, or head-specific policies~\cite{li2024snapkv,cai2024pyramidkv,xiao2024duoattention}. Structured sparse methods further exploit recurring attention patterns to accelerate long-context computation~\cite{jiang2024minference}. These approaches provide bounded memory or predictable computation, but may restrict access to positions excluded from the active cache or sparse pattern. In contrast, ReTopK preserves the complete KV cache without eviction, enabling query-dependent retrieval from the full history.

\paragraph{Query-aware sparse selection.}
Query-aware methods reduce attention cost by estimating relevant context from the current query. SparQ estimates key relevance from selected query dimensions, Quest prunes KV pages with query-dependent bounds, and Loki searches a low-rank key space~\cite{ribar2023sparq,tang2024quest,singhania2024loki}. RetrievalAttention indexes KV entries, Squeezed Attention retrieves key clusters, and SeerAttention predicts sparse structures with learned gates~\cite{liu2024retrievalattention,hooper2024squeezed,gao2024seerattention}. TokenSelect caches a selection for similar consecutive queries~\cite{wu2024tokenselect}. Recent methods reuse sparse selections across layers or decoding steps: Kascade and IndexCache share Top-$K$ indices across layers, while PRR overlaps temporal reuse with exact selection and repairs missed blocks~\cite{deshmukh2025kascade,bai2026indexcache,wang2026prr}. ReTopK instead retrieves multiple supports from a bounded per-head query cache and reranks their union with current-query scores, avoiding a full-history scan on reuse steps; fallback and refresh maintain reliability without training or KV eviction.

\section{Methodology}
\label{sec:method}

\subsection{Motivation: Query Similarity and Support Reuse}
\label{sec:motivation}

We begin by examining whether historical attention decisions can help identify the important positions for a new query. We consider one attention layer and one query head, and omit these indices throughout this section. Let $q_t$ denote the query at decoding step $t$, and let $\{k_i\}_{i=1}^{L_t}$ be the keys currently available in the KV cache. The Exact Top-$K$ support is
\begin{equation}
S_t = \operatorname{TopK}_{i \in \{1,\ldots,L_t\}} \left( \frac{q_t^\top k_i}{\sqrt{d}}, K \right),
\label{eq:exact_support}
\end{equation}
where $S_t$ contains the indices of the $K$ largest scores.

For two queries at positions $t$ and $j$, we measure their normalized support overlap as
\begin{equation}
\operatorname{Overlap}(t,j) = \frac{\left|S_t \cap S_j\right|}{K}.
\label{eq:support_overlap}
\end{equation}
Given the $R$ most similar preceding queries $\{j_1,\ldots,j_R\}$, we further define the union of their supports and its coverage of the current Exact Top-$K$ support as
\begin{equation}
U_t(R) = \bigcup_{r=1}^{R} S_{j_r}, \; \operatorname{Coverage}_t(R) = \frac{\left|S_t \cap U_t(R)\right|}{K}.
\label{eq:support_coverage}
\end{equation}

\begin{figure}[!t]
\centering
\includegraphics[width=\columnwidth]{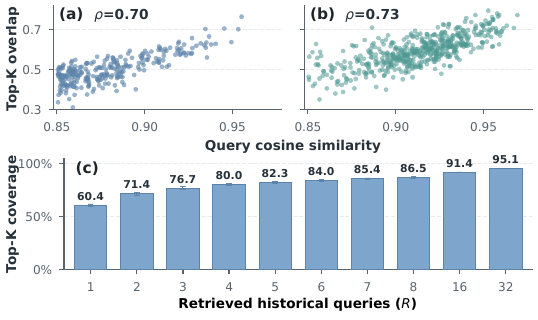}
\caption{Query similarity versus Exact Top-$K$ support overlap: (a) all
historical pairs; (b) each query's most similar predecessor; (c) coverage by
the union of $R$ retrieved supports.}
\label{fig:similarity_reuse}
\end{figure}

Figure~\ref{fig:similarity_reuse} shows that query cosine similarity is strongly correlated with Exact Top-$K$ support overlap. This relation holds both across historical query pairs and when each query is matched to its most similar preceding query. Although the support of a single historical query is incomplete, unioning supports from multiple similar queries substantially improves coverage.

\begin{figure}[!t]
\centering
\includegraphics[width=\columnwidth]{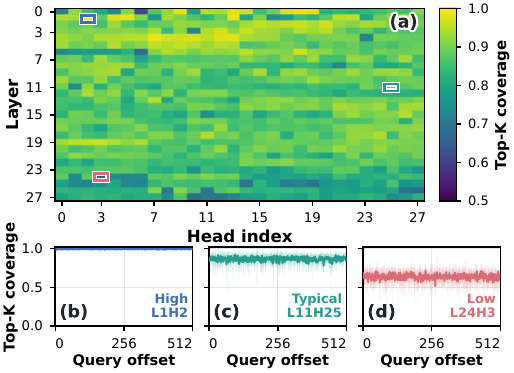}
\caption{Head-wise Exact Top-$K$ support coverage on Qwen2.5-7B at 16K
($K=512$, $R=8$): (a) layer--head means; (b--d) representative decoding
traces.}
\label{fig:headwise_coverage}
\end{figure}

Reuse quality is not uniform across attention heads. As shown in Figure~\ref{fig:headwise_coverage}, some heads achieve near-complete coverage from retrieved supports, whereas others exhibit substantially lower coverage. Within each head, however, coverage remains relatively stable over consecutive decoding steps. This head-dependent behavior motivates maintaining retrieval decisions independently for each query head. It also shows that reuse should not be forced universally: a practical method requires a mechanism to identify and recover from unreliable reuse.

These observations lead to three design principles. First, retrieval decisions should be stored together with their originating queries, so that query similarity can serve as a lightweight retrieval signal. Second, recalled supports should be treated as candidates rather than final selections, and should be reranked under the current query. Third, reuse must be paired with safeguards for low-confidence and stale retrieval decisions. ReTopK implements these principles through a bounded per-head query--support cache, exact reranking over recalled candidates, similarity-based fallback, and periodic refresh.

\subsection{Problem Setup and ReTopK Overview}
\label{sec:overview}

At decoding step $t$, let $q_t \in \mathbb{R}^{d}$ denote the current query, and let $\{(k_i,v_i)\}_{i=1}^{L_t}$ denote the KV cache. Exact Top-$K$ attention computes a score for every cached key,
\begin{equation}
e_{t,i} = \frac{q_t^\top k_i}{\sqrt{d}}, \qquad i \in \{1,\ldots,L_t\},
\label{eq:full_scores}
\end{equation}
selects the full-history support $S_t$ in Equation~\ref{eq:exact_support}, and performs Softmax--value aggregation over the selected entries. Although this aggregation involves only $K$ entries, obtaining $S_t$ still requires full-history QK scoring and global Top-$K$ selection.

On reuse steps, ReTopK replaces this full-history selection with a \emph{recall-before-rerank} procedure. It first retrieves historical supports associated with queries similar to $q_t$, then constructs a compact candidate set from these supports and a recent window. ReTopK computes exact current-query scores only for the candidate keys and selects the final Top-$K$ entries within this set. When reuse is unreliable, it reverts to Exact Top-$K$ attention.

\begin{figure*}[]
\centering
\includegraphics[width=\textwidth]{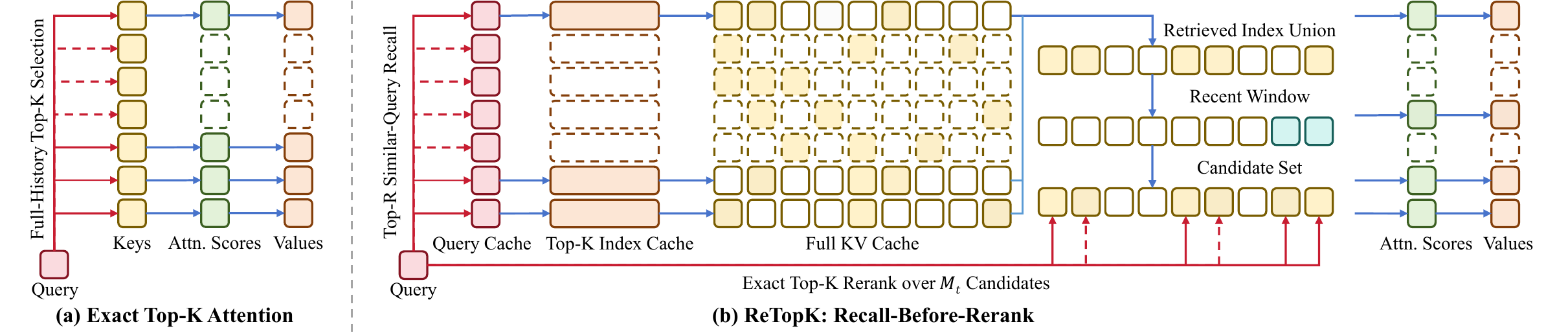}
\caption{(a) Full-history Exact Top-$K$ selection. (b) ReTopK recall,
candidate construction, and reranking. Dashed boxes and lines denote
unselected entries and connections.}
\label{fig:retopk_overview}
\end{figure*}

Figure~\ref{fig:retopk_overview} compares this procedure with Exact Top-$K$ attention. Exact Top-$K$ scores all $L_t$ keys at every step, whereas ReTopK searches only a bounded cache of historical queries and reranks $M_t \ll L_t$ candidate keys on the reuse path. Importantly, ReTopK reuses only historical \emph{index selections}; it does not reuse historical QK scores, attention probabilities, or attention outputs.

\subsection{Recall-Before-Rerank Selection}
\label{sec:selection}

\paragraph{Query--support cache.} For each query head, ReTopK maintains a bounded FIFO cache of capacity $C$:
\begin{equation}
\mathcal{B}_t = \left\{ \left(\bar q_c, \hat S_c\right) \right\}_{c=1}^{|\mathcal{B}_t|}, \qquad \bar q_c = \frac{q_c}{\|q_c\|_2},
\label{eq:cache}
\end{equation}
where $\bar q_c$ is the normalized query representation after positional encoding, and $\hat S_c$ is the support selected at the corresponding historical step. The cache is initialized from the final $C$ prompt tokens during prefill and updated after each decoding step, whether ReTopK takes the reuse or exact path. The cache adds bounded metadata to the original KV cache: it stores normalized queries and selected indices, while preserving the complete set of keys and values.

\paragraph{Similarity-guided recall.} For the normalized current query $\bar q_t = q_t / \|q_t\|_2$, ReTopK computes cosine similarities to all cached queries:
\begin{equation}
\rho_{t,c} = \bar q_t^\top \bar q_c, \qquad c \in \{1,\ldots,|\mathcal{B}_t|\}.
\label{eq:cache_similarity}
\end{equation}
It retrieves the indices of the $R$ most similar cache entries,
\begin{equation}
\mathcal{R}_t = \operatorname{TopR}_{c \in \{1,\ldots,|\mathcal{B}_t|\}} \left(\rho_{t,c}, R\right), \; \rho_t^{\max} = \max_c \rho_{t,c}.
\label{eq:topr_retrieval}
\end{equation}
Unlike Exact Top-$K$, this recall stage compares $q_t$ with only $C$ cached queries rather than all $L_t$ historical keys.

\paragraph{Candidate construction and reranking.} The supports associated with the retrieved cache entries are merged with a recent-token window,
\begin{equation}
\mathcal{L}_t = \left\{ \max(1, L_t-W+1), \ldots, L_t \right\},
\label{eq:local_window}
\end{equation}
where $W$ is the window size. The resulting candidate set is
\begin{equation}
\mathcal{A}_t = \operatorname{Unique} \left( \bigcup_{c \in \mathcal{R}_t} \hat S_c \;\cup\; \mathcal{L}_t \right), \qquad M_t = |\mathcal{A}_t|.
\label{eq:candidate_set}
\end{equation}
The recent window ensures that newly appended tokens remain eligible even if they do not appear in any retrieved support.

ReTopK then computes exact current-query QK scores only for candidate keys:
\begin{equation}
\hat e_{t,i} = \frac{q_t^\top k_i}{\sqrt{d}}, \qquad i \in \mathcal{A}_t,
\label{eq:candidate_scores}
\end{equation}
and selects the reranked support
\begin{equation}
\hat S_t = \operatorname{TopK}_{i \in \mathcal{A}_t} \left(\hat e_{t,i}, K\right).
\label{eq:reranked_support}
\end{equation}
Reranking maps the variable-size candidate set back to exactly $K$ indices,
preserving the fixed-size support stored in each cache entry. If
$S_t\subseteq\mathcal{A}_t$, this reranking step exactly recovers $S_t$.

Finally, ReTopK performs sparse attention over the reranked support:
\begin{equation}
\hat o_t = \sum_{i \in \hat S_t} \frac{\exp(\hat e_{t,i})}{\sum_{j \in \hat S_t}\exp(\hat e_{t,j})}v_i.
\label{eq:sparse_attention}
\end{equation}
Thus, historical supports are used only for candidate generation. All retained scores, normalized attention weights, and value aggregations are computed using the current query and the original KV cache.

\subsection{Reliable Reuse via Fallback and Refresh}
\label{sec:reliability}

Query similarity provides a useful but imperfect signal for reuse fidelity. ReTopK therefore takes the Exact Top-$K$ path whenever the maximum cached-query similarity falls below a threshold $\tau$, i.e., $\rho_t^{\max}<\tau$. This fallback avoids reliance on historical supports when no sufficiently related query is available.

Fallback alone does not fully address error accumulation, because supports produced on the reuse path are subsequently inserted into the cache. To periodically inject exact retrieval decisions, ReTopK performs an Exact Top-$K$ refresh every $T_r$ decoding steps; setting $T_r=0$ disables scheduled refresh. Let $t_{\mathrm{dec}}$ denote the decoding-step offset and $\mathcal{P}_t$ the selected path. Then
\begin{equation}
\mathcal{P}_t =
\begin{cases}
\textsc{Exact}, & T_r>0,\; t_{\mathrm{dec}}\in\{T_r,2T_r,\ldots\}, \\
\textsc{Exact}, & \rho_t^{\max}<\tau, \\
\textsc{Reuse}, & \text{otherwise}.
\end{cases}
\label{eq:path_decision}
\end{equation}
The two mechanisms are complementary: similarity fallback handles queries without reliable cached-query matches, whereas periodic refresh limits long-horizon drift in the online cache.

After either path, ReTopK caches $(\bar q_t,\hat S_t)$, with $\hat S_t=S_t$ on exact steps.

\subsection{Complexity and GPU Implementation}
\label{sec:complexity}

For Exact Top-$K$, full-history QK scoring requires $\mathcal{O}(L_t d)$ work, followed by global Top-$K$ selection over $L_t$ scores. In contrast, a ReTopK reuse step requires $\mathcal{O}(Cd)$ work for cache matching, $\mathcal{O}(RK+W)$ work to construct the raw candidate list, and $\mathcal{O}(M_t d)$ work for candidate QK scoring, followed by Top-$K$ selection over $M_t$ scores, where
\begin{equation}
M_t \leq RK + W \ll L_t.
\label{eq:candidate_bound}
\end{equation}
The subsequent sparse Softmax--value aggregation remains $\mathcal{O}(K d_v)$, where $d_v$ is the value-head dimension. Hence, with $C$, $R$, $K$, and $W$ fixed, a reuse step replaces both full-history QK scoring and global selection over $L_t$ entries with bounded cache lookup and candidate processing over $M_t$ entries. Its selector work therefore does not grow with context length. The additional metadata cost is $\mathcal{O}(Cd + CK)$ per head, while the complete KV cache remains unchanged.

Our GPU implementation maps the reuse path to three GPU stages. A single fused selector performs cache matching, Top-$R$ retrieval, and exact candidate sorting and deduplication. An indexed QK kernel directly scores the packed candidate keys without materializing a gathered key tensor. A final fused kernel performs candidate Top-$K$ selection, indexed Softmax--value aggregation, and cache update. Cache state and per-layer workspaces are preallocated and kept on the GPU. Exact, fallback, and refresh paths use matched full-history kernels, so the paths differ only in support discovery.

\begin{table*}[t]
\centering
{\small
\setlength{\tabcolsep}{2.0pt}
\renewcommand{\arraystretch}{1.10}
\begin{tabular}{l*{16}{c}}
\toprule
\multirow[c]{2}{*}{\textbf{Method}}
& \multicolumn{4}{c}{\textbf{PG19 PPL $\downarrow$}}
& \multicolumn{4}{c}{\textbf{Speedup vs. Exact Top-$K$ $\uparrow$}}
& \multicolumn{4}{c}{\textbf{NIAH: 3-task Avg. $\uparrow$}}
& \multicolumn{4}{c}{\textbf{LongBench $\uparrow$}} \\
\cmidrule(lr){2-5}\cmidrule(lr){6-9}\cmidrule(lr){10-13}\cmidrule(lr){14-17}
& \textbf{16K} & \textbf{32K} & \textbf{64K} & \textbf{128K}
& \textbf{16K} & \textbf{32K} & \textbf{64K} & \textbf{128K}
& \textbf{16K} & \textbf{32K} & \textbf{64K} & \textbf{128K}
& \textbf{2Wiki} & \textbf{Hotpot} & \textbf{MFQA} & \textbf{Avg.} \\
\midrule
\rowcolor{ReferenceRow}
Full Attention
& 8.78 & 8.76 & 8.82 & 11.62
& $2.08\times$ & $1.18\times$ & $0.95\times$ & $0.87\times$
& 99.5 & 97.3 & 98.5 & 93.2
& 55.1 & 59.8 & 49.8 & 54.9 \\
\rowcolor{ReferenceRow}
Exact Top-$K$
& 8.92 & 8.94 & 9.03 & 12.07
& $1.00\times$ & $1.00\times$ & $1.00\times$ & $1.00\times$
& 95.2 & 95.0 & 98.0 & 64.5
& 49.8 & 57.5 & 50.2 & 52.5 \\
\midrule
StreamingLLM
& 11.60 & 11.49 & 12.18 & 12.38
& \bestresult{$\boldsymbol{2.19\times}$} & \bestresult{$\boldsymbol{2.36\times}$}
& \bestresult{$\boldsymbol{3.61\times}$} & \bestresult{$\boldsymbol{6.39\times}$}
& 3.3 & 3.8 & 2.5 & 3.5
& 22.3 & 30.8 & 24.4 & 25.8 \\
Quest
& 10.55 & 17.05 & 18.17 & 36.23
& $1.07\times$ & $1.13\times$ & $1.75\times$
& \secondresult{$3.09\times$}
& 43.7 & 78.8 & 29.7 & 24.0
& 43.0 & 39.5 & 42.0 & 41.5 \\
SparQ
& 11.87 & 13.43 & 13.55 & 23.26
& $0.68\times$ & $0.51\times$ & $0.78\times$ & $1.39\times$
& 59.2 & 84.8 & 48.0 & 45.3
& \secondresult{47.4} & 52.9 & 48.8 & 49.7 \\
Loki ($r=64$)
& 61.01 & 33.28 & 110.08 & 199.32
& $0.57\times$ & $0.60\times$ & $0.93\times$ & $1.37\times$
& 4.7 & 0.0 & 0.0 & 0.0
& 40.4 & 39.5 & 46.9 & 42.3 \\
TokenSelect
& 10.25 & 10.60 & 11.12 & 17.71
& $0.67\times$ & $0.63\times$ & $0.97\times$ & $1.33\times$
& 65.8 & 83.3 & 36.7 & 44.8
& 47.3 & \secondresult{56.0}
& \secondresult{49.8} & \secondresult{51.0} \\
\rowcolor{OursRow}
\textbf{ReTopK ($\tau=0.85$)}
& \secondresult{8.98} & \secondresult{9.03}
& \secondresult{9.18} & \bestresult{12.13}
& \secondresult{$1.29\times$} & \secondresult{$1.38\times$}
& \secondresult{$2.03\times$} & $3.07\times$
& \secondresult{85.3} & \secondresult{89.3}
& \secondresult{93.2} & \secondresult{53.3}
& 47.0 & 54.5 & 48.9 & 50.1 \\
\rowcolor{OursRow}
\textbf{ReTopK ($\tau=0.90$)}
& \bestresult{8.95} & \bestresult{8.98}
& \bestresult{9.11} & \secondresult{12.20}
& $1.14\times$ & $1.23\times$ & $1.70\times$ & $2.25\times$
& \bestresult{96.3} & \bestresult{92.2}
& \bestresult{98.0} & \bestresult{75.2}
& \bestresult{48.4} & \bestresult{58.7}
& \bestresult{51.0} & \bestresult{52.7} \\
\bottomrule
\end{tabular}
}
\caption{Long-context performance and attention efficiency ($K=512$).
Speedup is relative to Exact Top-$K$. Gray and teal denote reference and
ReTopK rows, respectively; boldface and underlining mark the best and
second-best approximate results.}
\label{tab:main-quality-efficiency}
\end{table*}

\section{Experiments}

\subsection{Experimental Setup}
\paragraph{Models and benchmarks.}
We evaluate Qwen2.5-7B~\citep{qwen2024qwen25} on
PG19~\citep{rae2020compressive} and
Qwen2.5-7B-Instruct-1M~\citep{yang2025qwen251m} on RULER
NIAH~\citep{hsieh2024ruler} and
LongBench~\citep{bai2024longbench}. PG19 and NIAH are evaluated at
16K, 32K, 64K, and 128K contexts. For PG19, we report perplexity over a
512-token teacher-forced suffix for each of 64 fixed documents. For NIAH,
we report the macro-average of single-needle, multi-key, and multi-value
retrieval over 50 examples per task and context length. For LongBench, we
use the complete test splits of three subsets: 2WikiMQA, HotpotQA, and
MultiFieldQA, comprising 550 examples in total.

\paragraph{Baselines.}
We compare ReTopK with Full Attention, Exact Top-$K$, StreamingLLM, Quest,
SparQ, Loki, and TokenSelect~\citep{xiao2024streamingllm,tang2024quest,
ribar2023sparq,singhania2024loki,wu2024tokenselect}. Unless otherwise
specified, all sparse methods use a shared $K=512$ active-token budget.
StreamingLLM allocates this budget to 64 sink and 448 recent tokens,
whereas TokenSelect uses 64 initial, 416 selected, and 32 local tokens.
Other baseline configurations are provided in the appendix.

\paragraph{Implementation settings.}
Unless otherwise specified, ReTopK uses $C=W=32$, $R=4$, $\tau=0.85$,
and $T_r=128$. All experiments use BF16. Attention latency is measured
on a single NVIDIA L20 GPU using matched kernels, and speedup is reported
relative to Exact Top-$K$.

\subsection{Experimental Results}

\paragraph{Overall quality and efficiency.}
Table~\ref{tab:main-quality-efficiency} reports long-context performance and
attention efficiency. Across all context lengths, the two ReTopK settings rank
first and second among approximate methods in PG19 PPL and jointly attain the
best approximate results across all reported task metrics, while matching
or occasionally exceeding Exact Top-$K$. The default $\tau=0.85$ achieves a
$3.07\times$ speedup at 128K with a 0.50\% PPL increase, whereas the
quality-oriented $\tau=0.90$ maintains a $1.14$--$2.25\times$ speedup, leads
all approximate methods on NIAH at every length, and obtains a LongBench
average of 52.7, exceeding the 52.5 Exact Top-$K$ reference.

Speedup should be interpreted jointly with task performance. StreamingLLM
is fastest, but its fixed sink-and-recent window excludes evidence outside the
retained tokens, limiting its NIAH score to 2.5--3.8. TokenSelect reaches 51.0
on LongBench, yet its selection cache hits on only 34.7--35.7\% of NIAH steps,
restricting its speedup to $0.67$--$1.33\times$. Loki uses the shared $K=512$
budget and rank 64, substantially below the 12.5--25\% context-proportional
budgets used in its paper; its results therefore characterize a strict-budget
operating point. Paper-recommended configurations are compared in the
appendix.

\begin{table}[t]
\centering
{\small
\setlength{\tabcolsep}{1.0pt}
\renewcommand{\arraystretch}{1.07}
\begin{tabular}{@{}lccccccc@{}}
\toprule
\multirow[c]{2}{*}[-0.35ex]{\textbf{Model}} &
\multirow[c]{2}{*}[-0.35ex]{\textbf{Ctx.}} &
\multirow[c]{2}{*}[-0.35ex]{\textbf{$K$}} &
\multicolumn{2}{c}{\textbf{PPL}} &
\multirow[c]{2}{*}[-0.35ex]{\shortstack{\textbf{$\Delta$PPL}\\\textbf{(\%)}}} &
\multirow[c]{2}{*}[-0.35ex]{\shortstack{\textbf{Reuse}\\\textbf{(\%)}}} &
\multirow[c]{2}{*}[-0.35ex]{\shortstack{\textbf{Speedup}\\\textbf{vs. Exact}}} \\
\cmidrule(lr){4-5}
& & & \textbf{Exact} & \textbf{ReTopK} & & & \\
\midrule
\multirow[c]{4}{*}{Qwen2.5-7B}
& 16K  & 512  & 8.92  & 8.98  & $+0.70$ & 87.5 & $1.29\times$ \\
& 32K  & 512  & 8.94  & 9.03  & $+1.08$ & 87.7 & $1.38\times$ \\
& 64K  & 512  & 9.03  & 9.18  & $+1.60$ & 88.4 & $2.03\times$ \\
& 128K & 1024 & 11.72 & 12.05 & $+2.76$ & 89.2 & $1.76\times$ \\
\midrule
\multirow[c]{4}{*}{Llama-3.1-8B}
& 16K  & 512  & 7.93 & 8.06 & $+1.71$ & 86.3 & $1.29\times$ \\
& 32K  & 512  & 8.13 & 8.29 & $+2.00$ & 86.1 & $1.83\times$ \\
& 64K  & 512  & 8.15 & 8.37 & $+2.62$ & 85.9 & $2.66\times$ \\
& 128K & 1024 & 7.93 & 8.11 & $+2.28$ & 89.6 & $2.59\times$ \\
\midrule
\multirow[c]{4}{*}{Qwen2.5-14B}
& 16K  & 512  & 7.69 & 7.74 & $+0.55$ & 82.5 & $1.26\times$ \\
& 32K  & 512  & 7.72 & 7.74 & $+0.25$ & 82.8 & $1.81\times$ \\
& 64K  & 512  & 7.74 & 7.76 & $+0.21$ & 83.0 & $2.59\times$ \\
& 128K & 1024 & 8.38 & 8.34 & $-0.46$ & 86.3 & $2.52\times$ \\
\bottomrule
\end{tabular}
}
\caption{Cross-model PG19 results (64 documents; shared ReTopK settings;
reuse weighted over head--token pairs).}
\label{tab:cross-model}
\end{table}

\paragraph{Cross-model generalization.}
We apply the configuration selected on Qwen2.5-7B to
Llama-3.1-8B~\citep{grattafiori2024llama3} and
Qwen2.5-14B without model-specific tuning, using $C=W=32$, $R=4$,
$\tau=0.85$, and $T_r=128$, with $K=512$ at 16K--64K and $K=1024$ at 128K.
Across all model--context pairs, ReTopK
achieves 82.5--89.6\% reuse, PPL changes of $-0.46\%$ to $+2.76\%$ relative
to Exact Top-$K$, and speedups of $1.26$--$2.66\times$, demonstrating consistent
transfer across model families and scales.

\paragraph{Attention-support fidelity.}
ReTopK need not reproduce every index selected by Exact Top-$K$, provided that
the reranked support preserves high-mass positions and the resulting attention
output. We evaluate this criterion for both a representative head and all
layers and query heads.

\begin{figure}[t]
  \centering
  \includegraphics[width=\columnwidth]{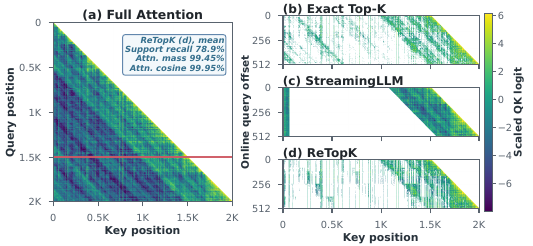}
  \caption{QK supports for Qwen2.5-7B layer 18, head 6 after a 1,536-token
  prefill ($K=512$; 512 decoding steps). White denotes unselected or masked
  entries.}
  \label{fig:qk-support-geometry}
\end{figure}

Figure~\ref{fig:qk-support-geometry} compares Full Attention, Exact Top-$K$,
StreamingLLM, and ReTopK for the representative head. ReTopK's long-range
support geometry aligns more closely with Exact Top-$K$ than StreamingLLM's
fixed sink-and-recent pattern. With 78.9\% Exact Top-$K$ support recall, ReTopK
retains 99.45\% of the attention mass and achieves 99.95\%
attention-distribution cosine, indicating that the mismatched indices receive
low normalized attention weights.

\begin{figure}[t]
  \centering
  \includegraphics[width=\columnwidth]{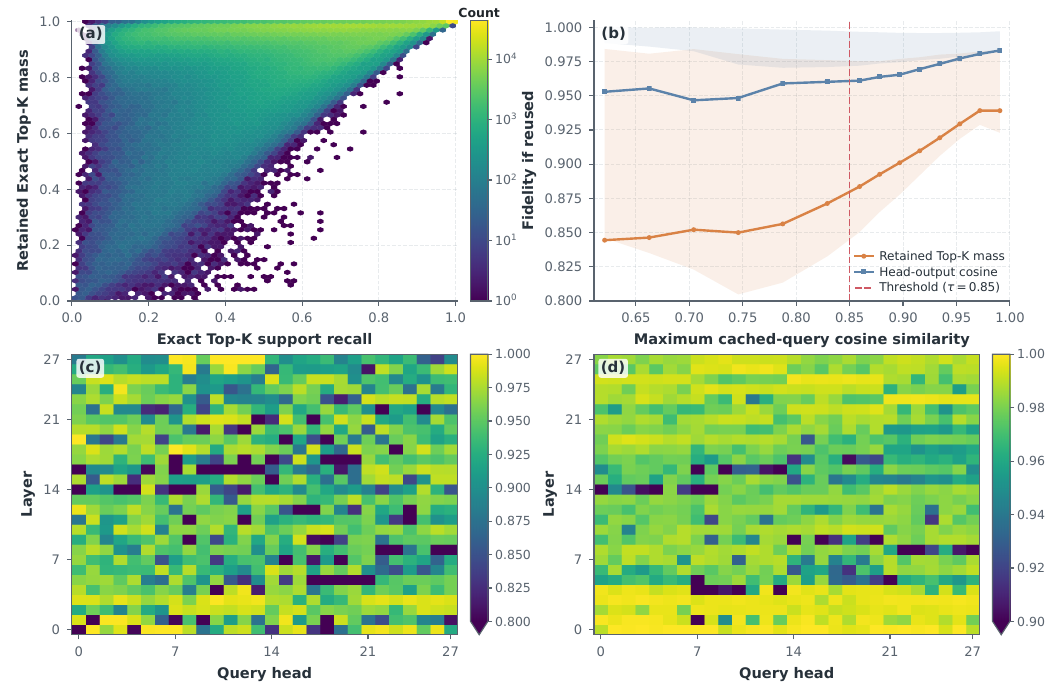}
  \caption{All-head ReTopK fidelity on Qwen2.5-7B for 4 PG19 documents
  (32K prefill, 512 decoding steps, $K=512$).}
  \label{fig:all-head-fidelity}
\end{figure}

Figure~\ref{fig:all-head-fidelity} extends the analysis to 1,605,632
head--token pairs across all 28 layers and 28 query heads. Retained mass and
head-output cosine average 92.4\% and 97.4\% over all pairs, and 91.6\% and
97.2\% on reuse-path pairs despite 56.9\% support recall. Both metrics
increase with cached-query similarity, supporting the similarity fallback at
$\tau=0.85$; the layer--head maps further show that this behavior is
distributed throughout the model.

\paragraph{Hyperparameter sensitivity and component ablation.}
\begin{figure}[t]
  \centering
  \includegraphics[width=\columnwidth]{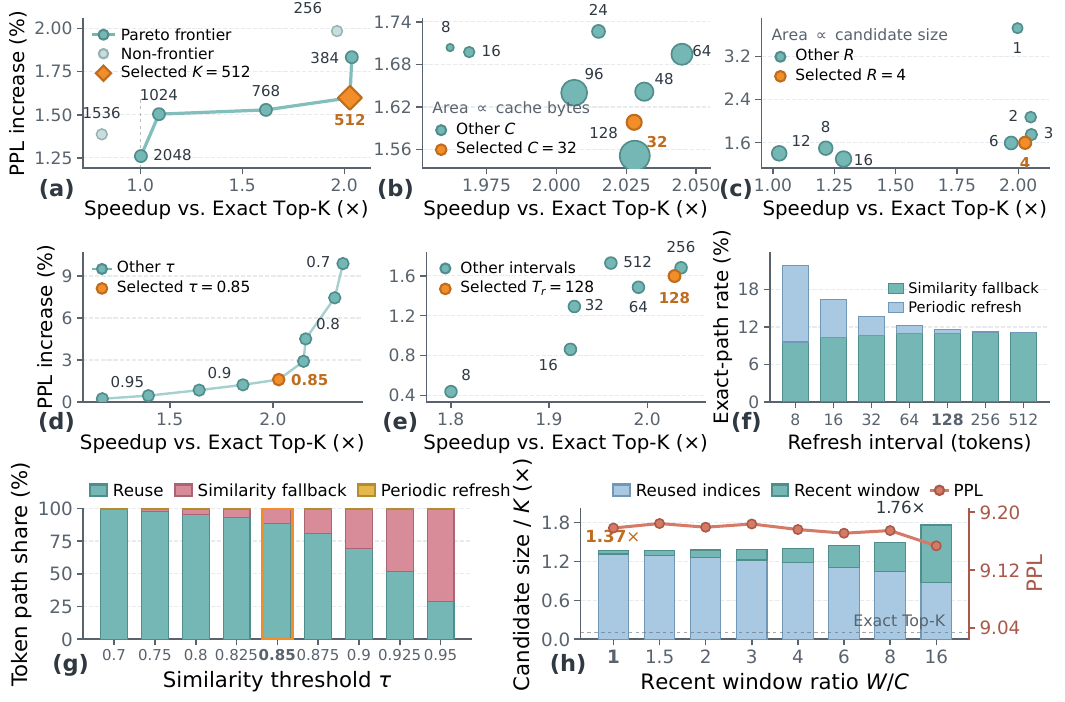}
  \caption{One-at-a-time ReTopK hyperparameter sensitivity at 64K; orange
  marks defaults.}
  \label{fig:hyperparameter-sensitivity}
\end{figure}

Figure~\ref{fig:hyperparameter-sensitivity} evaluates one-at-a-time sensitivity
around the default 64K configuration. In panels (a--e), the default $K=512$
lies on the Pareto frontier, with a
$2.03\times$ speedup and a 1.60\% PPL increase over Exact Top-$K$. Larger $K$
yields gradual PPL improvements at higher reranking cost, reducing speedup to
$1.09\times$ at $K=1024$. Performance is stable near the default $C$, while
gains from increasing $R$ diminish beyond $R=4$. Increasing $\tau$ trades
speed for quality by triggering more exact fallbacks. Refresh intervals
$T_r\geq128$ give similar short-horizon PPL, whereas shorter intervals improve
quality through more frequent Exact Top-$K$ steps.

Panels (f--h) decompose these trade-offs: increasing $T_r$ reduces the refresh
share, while increasing $\tau$ shifts decisions from reuse to similarity
fallback. Increasing $W/C$ from 1 to 16 enlarges the candidate set from
approximately 701 to 901 candidates with little PPL variation; $W=0$ is
examined in the component ablation.

\begin{table}[t]
\centering
\setlength{\tabcolsep}{1.2pt}
\renewcommand{\arraystretch}{1.10}
{
\small
\begin{tabularx}{\columnwidth}{@{}>{\raggedright\arraybackslash}X@{\hspace{1.0pt}}l r r r r@{}}
\toprule
\textbf{Ablation} & \textbf{Setting} & \textbf{PPL $\downarrow$} &
\textbf{Exact\,\%} & \textbf{Cand.} & \textbf{Spd.\,$\times$} \\
\midrule
\rowcolor{ReferenceRow}
Exact Top-$K$ & Reference & 9.034 & 100.0 & -- & 1.00 \\
\midrule
\rowcolor{OursRow}
ReTopK (default) & Default & 9.178 & 11.6 & 701 & 2.03 \\
Cached-query retrieval & Recent $R$ & 9.184 & 16.7 & 649 & 1.84 \\
Multi-query union & $R=1$ & 9.370 & 12.1 & 517 & 2.00 \\
Historical-query search & $C=1$ & 9.250 & 22.6 & 515 & 1.83 \\
Similarity fallback & $\tau=-1$ & 10.076 & 0.6 & 654 & 2.36 \\
Scheduled refresh & $T_r=0$ & 9.190 & 11.1 & 689 & 2.03 \\
Recent-token guard & $W=0$ & 716.0 & 8.0 & 621 & 2.15 \\
\bottomrule
\end{tabularx}
}
\caption{ReTopK component ablation at 64K ($K=512$).}
\label{tab:algorithm-component-ablation}
\end{table}

Table~\ref{tab:algorithm-component-ablation} isolates the components; Exact
and Cand. denote the exact-path rate and mean candidate count. Although
recent-$R$ retrieval raises the exact-path rate from 11.6\% to 16.7\%, it has
slightly higher PPL and lower speedup than similarity-guided retrieval.
Smaller $R$ or $C$ degrades PPL or speed, while removing
similarity fallback trades PPL for speed. Eliminating the recent-token guard
causes severe degradation. Scheduled refresh has little short-horizon effect,
motivating the long-horizon evaluation.

\paragraph{Long-horizon stability.}
After a 64K prefill, we evaluate 8K continuations on 8 PG19 documents with
$K=512$ and $T_r\in\{0,32,128,512\}$. Panel (a) uses shared teacher-forced
tokens; panel (b) compares autoregressive ReTopK with an Exact Top-$K$ shadow
on the same generated prefix.

\begin{figure}[t]
  \centering
  \includegraphics[width=\columnwidth]{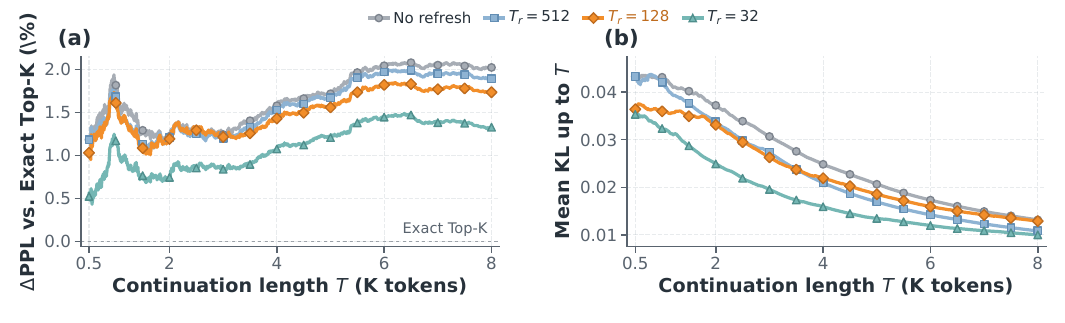}
  \caption{Long-horizon stability under (a) teacher forcing and
  (b) autoregressive generation.}
  \label{fig:long-horizon-stability}
\end{figure}

Panel (a) shows that quality error remains bounded through 8K: $\Delta$PPL is
within 2.08\% for all settings. More frequent refresh improves the 8K
endpoint from 2.02\% without refresh to 1.73\% at the default $T_r=128$ and
1.33\% at $T_r=32$, while the no-refresh setting remains stable. Panel (b)
yields 0.010--0.013 mean KL up to $T$ and 97.2--97.6\% Top-1 agreement under
autoregressive feedback, with no late-stage increase in error.

\paragraph{Kernel efficiency analysis.}
\begin{table}[t]
\centering
{
\small
\setlength{\tabcolsep}{1.2pt}
\renewcommand{\arraystretch}{1.06}
\begin{tabularx}{\columnwidth}{@{}l@{\hspace{4pt}}>{\raggedright\arraybackslash}Xrrrr@{}}
\toprule
& & \multicolumn{2}{c}{\textbf{64K}} &
    \multicolumn{2}{c}{\textbf{128K}} \\
\cmidrule(lr){3-4}\cmidrule(l){5-6}
\textbf{Level} & \textbf{Optimization stage}
& \textbf{Lat.} & \textbf{vs. L1}
& \textbf{Lat.} & \textbf{vs. L1} \\
\midrule
L1 & Separate-kernel baseline
& 6.630 & $1.00\times$ & 7.345 & $1.00\times$ \\
L2 & Candidate discovery and scoring
& 4.566 & $1.45\times$ & 5.230 & $1.40\times$ \\
L3 & Candidate selection and output
& 3.831 & $1.73\times$ & 4.486 & $1.64\times$ \\
L4 & Persistent workspaces
& 3.271 & $2.03\times$ & 3.886 & $1.89\times$ \\
\rowcolor{OursRow}
\textbf{L5} & \textbf{Candidate-to-output fusion}
& \textbf{2.577} & $\boldsymbol{2.57\times}$
& \textbf{3.207} & $\boldsymbol{2.29\times}$ \\
\bottomrule
\end{tabularx}
}
\caption{ReTopK kernel-fusion stages at 64K and 128K ($K=512$; ms/token
over 28 layers).}
\label{tab:fusion-ablation}
\end{table}

Table~\ref{tab:fusion-ablation} reports five cumulative stages under a fixed
input and path mixture. L1 separates retrieval and sparse output; L2 fuses
cache search with support union and adds indexed candidate scoring; L3 adds
exact sorting and deduplication while fusing selection, sparse output, and
cache update; L4 reuses per-layer workspaces; and L5 fuses candidate
construction with output.
Relative to L1, L5 achieves $2.57\times$ and $2.29\times$ speedups at 64K and
128K, respectively.

\begin{figure}[t]
  \centering
  \includegraphics[width=\columnwidth]{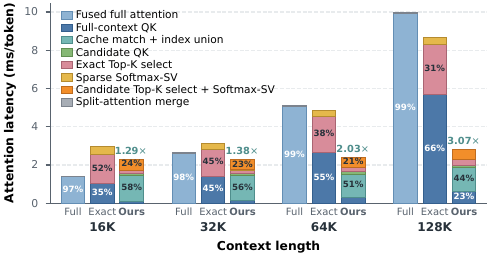}
  \caption{Attention latency composition across context lengths ($K=512$);
  labels above ReTopK denote speedup over Exact Top-$K$.}
  \label{fig:operator-breakdown}
\end{figure}

Figure~\ref{fig:operator-breakdown} decomposes attention latency across context
lengths. Full Attention remains scan-dominated at 97--99\%, while QK scoring
and Top-$K$ selection account for an increasing 87--97\% of Exact
Top-$K$ latency, driven primarily by QK and consistent with
Figure~\ref{fig:teaser}.
ReTopK's reuse-path cost remains nearly constant; its growth arises from
full-history QK on fallback and refresh steps. This growth does not arise from
the path mix: reuse rises from 87.5\% to 88.9\%, while low-similarity fallback
falls from 11.9\% to 10.5\%. Instead, the per-exact-path cost grows with
context length, increasing ReTopK's speedup over Exact Top-$K$ from
$1.29\times$ to $3.07\times$.

\paragraph{End-to-end long-context scaling.}
We measure matched end-to-end decoding on Qwen2.5-7B-Instruct-1M, using the
same GPU layout across methods and a fixed ReTopK path mixture across lengths.
Timings exclude prefill, setup, and compilation.

\begin{table}[t]
\centering
{
\small
\setlength{\tabcolsep}{1.4pt}
\renewcommand{\arraystretch}{1.10}
\begin{tabular}{@{}cc@{\hspace{1.5pt}}rrrcc@{}}
\toprule
\textbf{Ctx.} &
\textbf{GPU} &
\textbf{Full} &
\textbf{Exact} &
\cellcolor{OursRow}\textbf{ReTopK} &
\textbf{Non-attn.} &
\textbf{Spd.} \\
\midrule
128K & 1 & 30.30 & 28.88 & \cellcolor{OursRow}25.63 & 70.1\% & $\boldsymbol{1.13\times}$ \\
256K & 1 & 40.26 & 36.62 & \cellcolor{OursRow}27.74 & 55.3\% & $\boldsymbol{1.32\times}$ \\
512K & 2 & 60.51 & 51.48 & \cellcolor{OursRow}31.89 & 39.4\% & $\boldsymbol{1.61\times}$ \\
1M   & 2 & 96.20 & 88.81 & \cellcolor{OursRow}38.59 & 22.9\% & $\boldsymbol{2.30\times}$ \\
\addlinespace[1pt]
2M & 4 & 176.22 & 157.28 & \cellcolor{OursRow}52.72 & 13.0\% & $\boldsymbol{2.98\times}$ \\
3M & 8 & 247.85 & 224.00 & \cellcolor{OursRow}67.02 & 9.2\% & $\boldsymbol{3.34\times}$ \\
4M & 8 & 325.11 & 290.46 & \cellcolor{OursRow}81.51 & 7.1\% & $\boldsymbol{3.56\times}$ \\
5M & 8 & 399.81 & 357.10 & \cellcolor{OursRow}95.71 & 5.8\% & $\boldsymbol{3.73\times}$ \\
\bottomrule
\end{tabular}
}
\caption{End-to-end decoding latency (ms/token) across context lengths
($K=512$).}
\label{tab:end-to-end-scaling}
\end{table}

Table~\ref{tab:end-to-end-scaling} separates attention gains from shared
decoder work. Non-attn. denotes shared latency as a fraction of Exact Top-$K$
latency, and Spd. is relative to Exact Top-$K$. As the Non-attn. share decreases
from 70.1\% at 128K to 22.9\% at 1M and 5.8\% at 5M, ReTopK's speedup increases
from $1.13\times$ to $2.30\times$ and $3.73\times$, respectively. The
end-to-end implementation evaluates the full $RK+W=2080$ candidate capacity
without truncation.

\section{Conclusion}
We introduced ReTopK, a training-free recall-before-rerank method that
constructs candidate sets from cached query--support pairs, reranks them with
current-query scores, and uses similarity fallback and periodic refresh for
reliability. Across long-context language modeling and retrieval tasks, ReTopK
preserves the quality of Exact Top-$K$ while accelerating dynamic Top-$K$
attention by $3.07\times$ at 128K with a 0.50\% PPL increase. These results show
that retrieval-decision reuse reduces full-history index-discovery cost without
evicting KV entries.

\bibliography{references}

\end{document}